# SWTF: Sparse Weighted Temporal Fusion for Drone-Based Activity Recognition

Santosh Kumar Yadav
*College of Science and Engineering*
*National University of Ireland*
Galway, H91TK33, Ireland
santosh.yadav@ieee.org

Esha Pahwa
*Department of CSIS*
*Birla Institute of Technology and Science*
Pilani - 333031, India
f20180675@pilani.bits-pilani.ac.in

Achleshwar Luthra
*Department of CSIS*
*Birla Institute of Technology and Science*
Pilani - 333031, India
f20180401@pilani.bits-pilani.ac.in

Kamlesh Tiwari
*Department of CSIS*
*Birla Institute of Technology and Science*
Pilani - 333031, India
kamlesh.tiwari@pilani.bits-pilani.ac.in

Hari Mohan Pandey
*Department of Computer Science*
*Bournemouth University*
Poole BH12 5BB, United Kingdom
profharimohanpandey@gmail.com

Peter Corcoran
*College of Science and Engineering*
*National University of Ireland*
Galway, H91TK33, Ireland
peter.corcoran@nuigalway.ie

*Abstract*—Drone-camera based human activity recognition (HAR) has received significant attention from the computer vision research community in the past few years. A robust and efficient HAR system has a pivotal role in fields like video surveillance, crowd behavior analysis, sports analysis, and human-computer interaction. What makes it challenging are the complex poses, understanding different viewpoints, and the environmental scenarios where the action is taking place. To address such complexities, in this paper, we propose a novel Sparse Weighted Temporal Fusion (SWTF) module to utilize sparsely sampled video frames for obtaining global weighted temporal fusion outcome. The proposed SWTF is divided into two components. First, a temporal segment network that sparsely samples a given set of frames. Second, weighted temporal fusion, that incorporates a fusion of feature maps derived from optical flow, with raw RGB images. This is followed by base-network, which comprises a convolutional neural network module along with fully connected layers that provide us with activity recognition. The SWTF network can be used as a plug-in module to the existing deep CNN architectures, for optimizing them to learn temporal information by eliminating the need for a separate temporal stream. It has been evaluated on three publicly available benchmark datasets, namely Okutama, MOD20, and Drone-Action. The proposed model has received an accuracy of 72.76%, 92.56%, and 78.86% on the respective datasets thereby surpassing the previous state-of-the-art performances by a significant margin.

## Introduction

Human Activity Recognition (HAR) has a variety of applications, including video surveillance, security, crowd behavior analysis, and human-computer interaction, among others. It consists of two essential subtasks: classification and localization. Localization refers to the location of the action within a video scene, whereas classification determines what the person is doing. Our research encompasses the classification of actions using an efficient method to capture both temporal and spatial data in a single-stream model, as well as the classification of actions involving the handling of small objects.

Human Action Recognition presents significant challenges that must be addressed. Modern networks, such as 3D ConvNets, incur enormous computational expenses. Training a 3D CNN backbone requires a considerable amount of time, delaying the search for an optimal architecture and causing it to overfit the training data. Leaving aside the duration of model training, it should also be noted that HAR lacks a fixed dictionary of human activities. This can lead to diversity within the class. To address this, accurate and differentiating features need to be developed. Videos captured from a distance, such as those captured by video surveillance cameras, are also a potential impediment to identifying the correct action because they cannot provide high-quality, clear images of the person. Performance recognition also differs from the type of camera used. Real-time videos are visually dense and contain varying levels of brightness, making it difficult to see actions in complex situations. In addition to background activities performed by neighboring objects, variations in scale, viewpoint, and partial occlusion also influence the model's results.

Owing to the recent surge in the literature on this topic, a large number of studies have been conducted on human activity recognition [1]–[3]. In recent years, multimodal methods for HAR have gained popularity. Consequently, for a model to be successfully implemented, it must be efficient and accurate. Due to these reasons, works such as Persistent Appearance Network (PAN) [4] and Temporal Shift Module (TSM) [5], which can be utilized with both 3D and 2D CNNs, were brought about. Various types of CNN architectures *e.g.* [6] and [7] implement two-stream networks of 2D-CNNs whereas works like [8] are excellent example of 3D-CNN networks being both efficient and precise in their prediction. To learn long-range information from videos, Wang *et al.* proposed Temporal Segment Networks (TSN) [9] which uses segmented samples prior to feeding them to CNN architecture. Works such as Temporal Relation Network [10], Temporal Spatial Mapping [11], VLAD3 [12] and ActionVLAD [13], utilize this approach to deploy an efficient model.

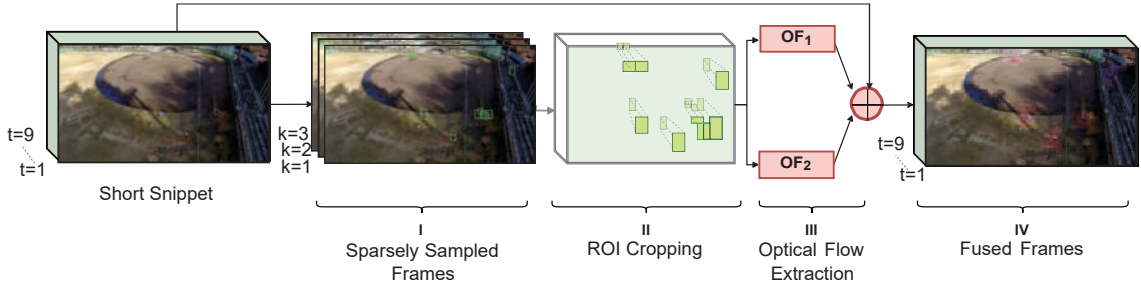

Fig. 1. Detailed description of the SWTF module. I) shows sparsely sampled frames obtained from a short snippet using a segment-based sampling technique. II) depicts how ROI(s) is(are) cropped across frames. III) and IV) illustrate optical flow extraction and their fusion with raw RGB frames respectively. Here values taken for *t* and *k* are for the demonstration purpose.

The current state-of-the-art for the Drone-Action dataset [14] uses a pose-stream that heavily relies on accurate joint estimations. The state-of-the-art on the MOD20 dataset [15] employs a two-stream strategy and relies on motion-CNN for accuracy. The state-of-the-art on the Okutama-Action dataset [16] uses features computed by 3D convolution neural networks in addition to a new set of features computed by the Binary Volume Comparison (BVC) layer, which consists of three components: a 3D-Conv layer with 12 non-trainable (*i.e.*, fixed) filters, a non-linear function, and a set of learnable weights. Features from both streams, 3D CNNs and BVC layer, are merged and sent to the Capsule Network for final activity prediction. Our computationally efficient SWTF (Sparse Weighted Temporal Fusion) model advances the state-of-the-art on the Drone-Action, MOD20, and Okutama-Action datasets without separately employing a pose-stream or temporal stream.

The previous works have mainly relied on a separate temporal stream to learn information available across a video. Our work tries to eliminate the need of using a temporal stream by introducing a novel approach to fusing raw RGB frames with the optical flow in an efficient way. We demonstrate in our study how our plug-in module can help reduce computation by a huge margin thus making drone-camera-based HAR more practical, fast, and coherent. Our method uses segment-based sampling to include global temporal information with a minimum number of frames. Our SWTF module can also boost the performance of existing approaches since it does not require knowledge about the internal details of the architecture such as activation functions, hidden layers, *etc.* and can be easily included in any method. The approach is easy to implement thus supporting faster experimentation. Other than that, the module can also act as a teacher network (existing network being the student network), optimize the existing network to learn temporal information, and then it can be removed at the time of performance evaluation. The major contributions of this paper have been listed below:

- We introduce Sparse Temporal Sampling before the Weighted Temporal Fusion module to bring attention to the location where the change is taking place, with significantly lesser computation.
- We incorporate Region Of Interest (ROI) Cropping in the Weighted Temporal Fusion module to deal with the extremely small size (as shown in Fig. 1) of human subjects. This helps us to recognize human activities from the high altitude of drone camera videos.
- The proposed SWTF module can act as a plug-in module.
- We perform extensive experimental analysis on three publicly available benchmark datasets, *i.e.*, Okutama dataset [17], MOD20 dataset [15], and Drone-Action dataset [18].

### PROPOSED METHODOLOGY

In this section, we give a detailed description of the individual components used in our model architecture. Then we explain how we have compiled those components to perform effective drone-camera based human action recognition. Our model takes a short snippet of video frames that undergoes necessary preprocessing. Then we select K frames using sparse temporal sampling. We use OpenCV to obtain optical flow for K frames and then fuse optical flow with the RGB frames using the Weighted Temporal Fusion module. After that, we extract features from fused frames (RGB and Optical Flow) using our backbone network, *i.e.*, Inception-v3 with Batch-Normalization. This is followed by the ROIAligning module which is used to concatenate features corresponding to our subjects. These features are further flattened and passed to fully connected layers which are followed by max-pooling resulting in individual-level action classification.

### *Data Preprocessing*

We have used three different datasets to verify the performance of our approach. Each dataset differs from the others in terms of actions, number of frames, frame rate, resolution of cameras used to record the videos, environment, camera motion, and even annotations. While the MOD20 dataset only comes with ground truth action labels, the Okutama dataset provides ground truth bounding boxes as well. The Drone-Action dataset goes one step ahead and provides ground truth pose annotations along with bounding boxes and frames. We have predicted bounding boxes separately for MOD20 as the intermediate layers of our novel WTF module rely on bounding box coordinates. We utilize the joint annotations provided with the Drone-Action dataset in a separate pose stream for a fair comparison with previous state-of-the-art methods. We use data augmentation techniques, such as random cropping and horizontal flip to prevent our model from adversarial examples. We resize our images while maintaining the aspect ratio. All the images are rescaled before being fed to the model, and bounding box coordinates are normalized as well.

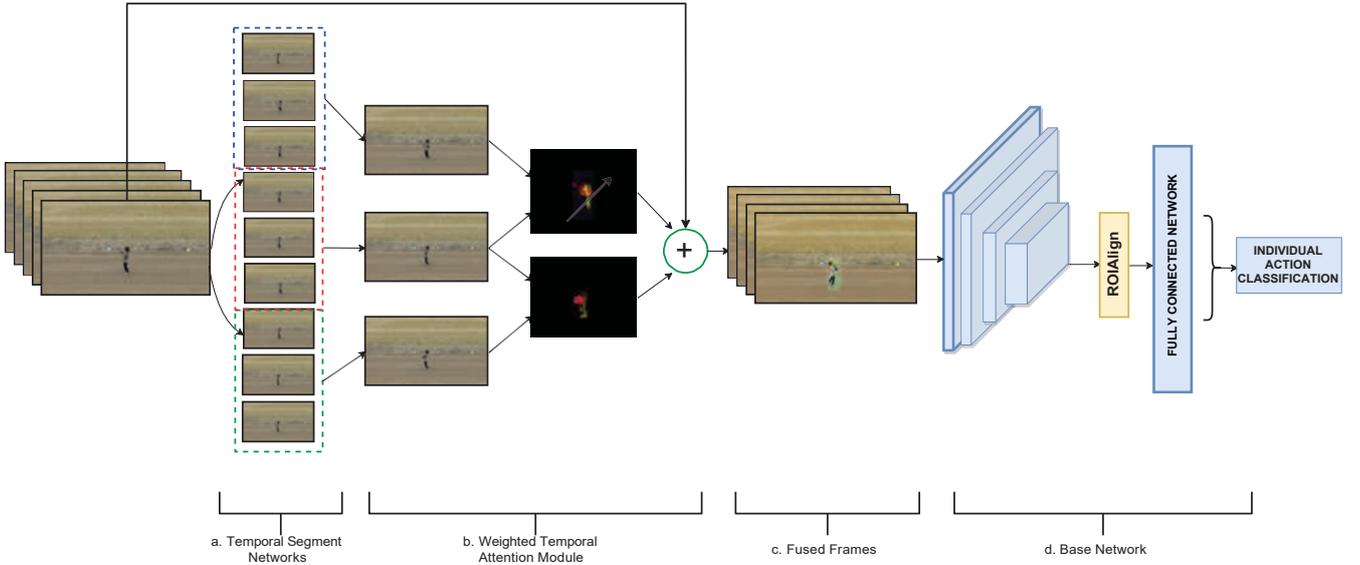

Fig. 2. Block diagram of the SWTF Network. (a). A clip from the extracted frames is sparsely sampled and segmented into equal halves. (b). Random frames is chosen out of those segments and using the Weighted Temporal Fusion module, we derive optical flow feature maps. (c). These maps are fused with the original frames. (d). The fused frames are fed into the base network which provides us with the activity classification results.

$$\text{Final Image} = \frac{\text{Original Image}}{255.0} - 0.5 \times 2.0 \quad (1)$$

*Temporal Segment Network*

As discussed in [this paper], dense sampling causes 2D ConvNets to overfit the training dataset as the frames are densely recorded in a video, and the content changes relatively slowly resulting in limited temporal information. Instead of using all the frames, we adopt a computationally efficient method [9] which helps to speed up the training process. We use sparse and global sampling techniques constructed using segment-based sampling to extract information across the entire snippet with a very less number of frames. The segment count is fixed thus guaranteeing that the computational cost will be constant throughout all the snippets.

Given a short snippet $S$ whose shape is $(T, C, H, W)$ where $T$ is no. of frames in a snippet, $C$ is a channel (eg: 3 for RGB), $H$ and $W$ are height and width respectively.

$$S = \{f_1, \ldots, f_T\}; \quad \text{where } f_i \in \mathbb{R}^{(C \times H \times W)} \quad \forall i \in [1, T] \quad (2)$$

where $f_i$ denotes $i_{th}$ frame in the snippet.

We divide it into $K$ segments $\{SGM_1, \ldots, SGM_K\}$ of equal durations and select one frame from each segment based on random sampling, as demonstrated in Fig. 3.

$$SGM_i = \{f_{((i-1) \cdot k)+1}, \ldots, f_{(i \cdot k)}\}; \quad \forall (i \cdot k) \leq T, i \leq K \quad (3)$$

We randomly select one frame F from each segment which implies:

$$F_i \in SGM_i \quad \forall i \in [1, K] \quad (4)$$

The use of a sparse sampling strategy reduces computational complexity dramatically and prevents overfitting which would have otherwise occurred due to a limited number of frames. Thus, it provides us with an efficient video-level framework that is capable of capturing long-range temporal structures.

*Weighted Temporal Fusion*

We have developed our novel Weighted Temporal Fusion module inspired by extensive research on applications of optical flow in the past years. This module takes sparsely sampled frames from Temporal Segment Network as input whose shape = $(K, C, H, W)$. It captures the motion of specific parts (Fig. 1) of input relevant to the task in hand and the resultant feature maps automatically lead to a sizable improvement in accuracy over baseline architectures.

Let $O(x, y)$ denote optical flow between $x$ and $y$, x and y being two frames:

$$OF_i = O(F_i, F_{i+1}) \quad (5)$$

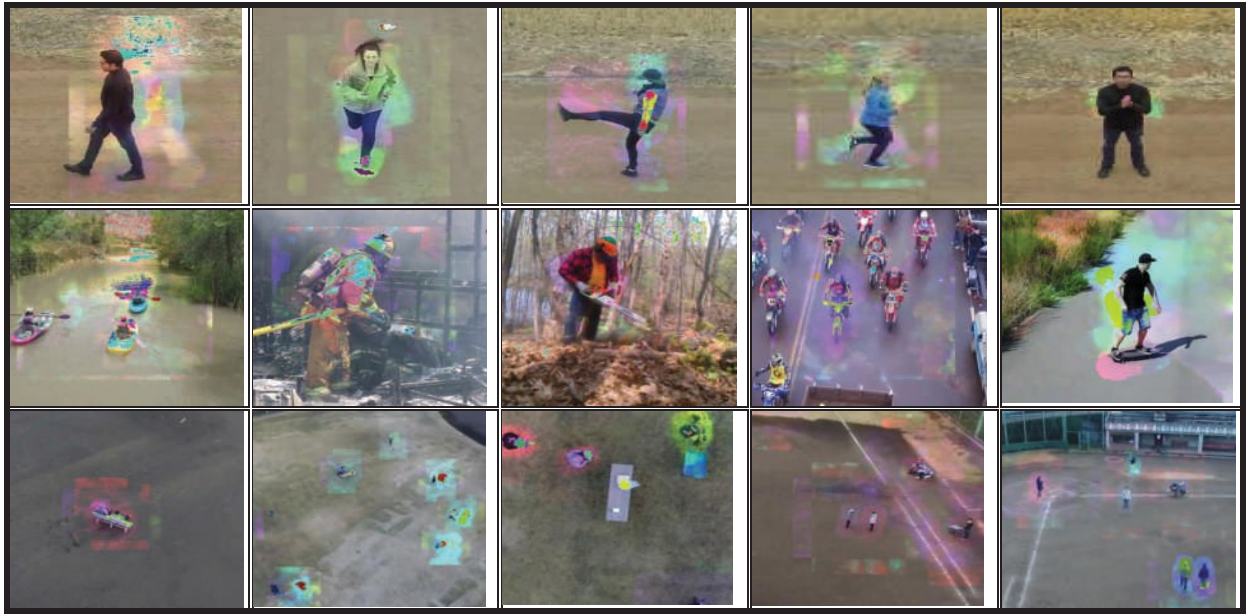

Fig. 3. The given set of images depicts the action of weighted temporal fusion of some selected frames. The first row contains examples from the Drone-Action dataset, of classes namely 'walking-sideways', 'running', 'kicking', 'running-sideways', and 'clapping' respectively. The second row shows examples from the MOD20 dataset of classes 'kayaking', 'fire-fighting', 'chainsawing-trees', 'motorbiking', and 'skateboarding' respectively. The third row shows examples from the Okutama dataset of various scenes where people can be seen sitting on a bench and interacting (first column); walking, standing (second column); carrying objects and interacting (third column); and standing (fourth and fifth column).

Let $x_F$ denote the weighted temporal fusion of snippet S:

$$x_F = \sum_{i=1}^{K-1} W_i \cdot OF_i \quad (6)$$

We perform element-wise multiplication of $x_F$ with all the T frames in our snippet S such that:

$$x_t := x_F \ \text{\textcircled{S}}\ x_t; \quad \forall t \in [1,T] \quad (7)$$

In our work, we take a clip of T=15 frames. This snippet is divided into 3 segments of 5 frames each, and a random frame is sampled from each of these segments. Thus, we now get 3 frames - ($F_1$, $F_2$, $F_3$) from which optical flow is calculated: $OF_1 = O(F_1, F_2)$, $OF_2 = O(F_2, F_3)$. Each of these optical flow values is multiplied with $W_i$ as 0.033 and summation takes place ($x_F$). This value is then multiplied by each of the 15 frames.

Weighted temporal fusion module is a simple yet powerful module to incorporate optical flow in action recognition. Our novel formulation is simple to implement and can be seen as an extension of "weighted average pooling". We add bounding box coordinates in the intermediate layers which encourage our novel module to look for relevant actions. We finally get an output of the shape - (1, C, H, W), which are further fused with RGB frames as shown in Fig. 2. This approach helps us in reasoning for the long-term temporal relations even by looking at a single frame. We finally combine appearances from static RGB images and motion inferred by the module to perform action recognition [19]. Examples of fused frames are given in Fig. 3.

*Backbone Network*

Most of the previous work including state-of-the-art algorithms incorporate two-stream ConvNets in their architecture to deal with appearance and motion separately. But the question is can we classify activities using a single stream of CNNs? In our approach, we have tried to deal with this issue of extra computation by merging optical flow features with the static images and used a single-stream of Inception-v3 [20] to predict individual-level actions. Pre-training the backbone on large-scale image recognition datasets, such as ImageNet [21], has turned out to be an effective solution when the target dataset does not have enough training samples [22]. We use Inception-v3 [20] with Batch Normalization pre-trained on ImageNet, as a backbone network, due to its balance between accuracy and efficiency. Our model falls under the risk of overfitting due to a limited number of training samples. To prevent this, we have relied on various regularization techniques. Batch Normalization is able to deal with the problem of covariate shift by estimating the activation mean and variance within each batch to normalize these activation values. This also helps in faster convergence. Further, we add dropout layers between our last fully connected layers having a dropout ratio of 0.3 before the global pooling layer. We use Adam optimizer with weight decay parameter set to $1e^{-4}$ which adds L2 norm regularization. These techniques prevent the high risk of overfitting and help in the generalization of our network.

TABLE I
SUMMARY OF DATASETS USED

| Dataset Name | Classes | #Clips | Duration | FPS |
|---|---|---|---|---|
| Okutama (2017) | 12 | 43 | 60.00 s | 30.00 |
| DroneAction (2019) | 13 | 240 | 11.15 s | 25.00 |
| MOD20 (2020) | 20 | 2324 | 7.40 s | 29.97 |

TABLE II
COMPARISON WITH THE STATE-OF-THE-ART RESULTS ON THE OKUTAMA DATASET. BLUE REPRESENTS THE PREVIOUS STATE-OF-THE-ART. RED DENOTES THE BEST RESULTS.

| S.no. | | Method | Backbone | Accuracy |
|---|---|---|---|---|
| Past Work | | AARN [23] [16] | C-RPN + YOLOv3-tiny | 33.75% |
| | | Lite ECO [24] [16] | BN-Inception + 3D-Resnet-18 | 36.25% |
| | | I3D(RGB) [25] [16] | 3D CNN backbone | 38.12% |
| | | 3DCapsNet-DR [26] [16] | 3D CNN + Capsule | 39.37% |
| | | 3DCapsNet-EM [26] [16] | 3D CNN + Capsule | 41.87% |
| | | DroneCaps [16] | 3D CNN + BVC + Capsule | 47.50% |
| Ours | | BaseNet | Inception-v3 | 61.34% |
| | | **SWTF** | **Weighted Temporal Fusion + Inception-v3** | **72.76%** |

## EXPERIMENTAL RESULTS

This section presents the experimental results of the proposed model on three datasets. The complete architecture of the proposed network used is given in Fig. 2. The summary of the dataset used is provided in TABLE I.

### Experimental Settings

Training of the model was carried out for 80 epochs for each dataset, on a system with an Intel Xeon processor, 12GB VRAM, and Nvidia Titan XP GPU. The model was compiled using the Pytorch backend. All the frames collected from the video datasets were first resized into a shape of 420 720 and normalized. Along with this, they were grouped into a batch size B=2 while taking frames T=15 at a time. The resulting data had a shape of (B, T, H, W, C) where H, W, and C denote height, width, and the number of channels of the frame, respectively.

Using the Inceptionv3 backbone, the feature maps obtained were processed in the ROIAlign function, having a crop size of 5, to get our desired region of interest. Therefore, the result was flattened and fed into a fully connected (FC) layer having $M = 512$ units, followed by a dropout layer, with the dropout ratio being 0.3, and a batch normalization layer. The output of the FC block was passed to the classifier which gave us the resulting probability. The train-to-test split ratio was held constant at 80:20 for all datasets. Adam optimizer with an initial learning rate of $10^{-5}$, $\beta_1$ and $\beta_2$ with a value of 0.9 and 0.999, and a weight decay of $10^{-4}$ was found to be the most suitable optimizer for the given task of action recognition as well as to prevent overfitting. The learning rate scheduler was utilized to decrease the learning rate by one-tenth of its value after every 30 epochs. The one-hot encoded targets and predictions were fed into Binary Cross Entropy Logits Loss owing to its satisfactory usability to process softmax outputs of the last layer of the model.

### Performance Evaluation

In this subsection, we discuss the evaluation results obtained on the ablation studies performed on Okutama, MOD20, and Drone Action datasets. TABLE II, III and IV summarize the results of the respective datasets with the mentioned backbone that is utilized.

Initially, we start our experiments using an Inception-v3 module to capture the spatial features of the original images without the weighted temporal fusion module. This helps us understand the critical and influential effect of the Weighted Temporal Fusion module. Training the model using a basic backbone with RoiAlign, fully connected, batch-normalization, and dropout layers resulted in 61.34% accuracy for the Okutama dataset. This alone can be seen as outperforming the previous state-of-the-art values achieved using different backbones. To further improve the outcome, the proposed Weighted Temporal Fusion module is added, and the backbone is fed with "fused" frames instead of original ones, resulting in an overall accuracy of 72.76%, marking an increase of 11.42% from the basenet architecture and 25.26% from the previous state of the art. Using the optical flow backdrop, the network can specifically focus on the region where the action is taking place, and disregard the background which may contain noise.

Similarly for MOD20 dataset, which consists of a diverse range of action classes, each significantly different from the other. It contains complex outdoor scenarios. That being said, our basenet model was able to achieve a higher accuracy of 90.03% as compared to the previous state-of-the-art value which was 74% [15]. After integrating the Weighted Temporal Fusion module, a slight increase of 2.56% was obtained.

The Drone Action dataset contained various action classes which were similar to one another. For example, jogging from the front, back, and sideways was similar to running front, back, and sideways. It was critical to exactly locate the joint positions in order to determine which action was being performed. Hence, without the pose annotations, results were obtained from our simple basenet: 62.79% and integrated Weighted Temporal Fusion module with basenet: 71.79%. Training the model along with pose joints in a separate stream led us to achieve greater results than the previous state-of-the-art, marking the increase by 2.84%.

### Discussion and Comparison

Our approach outperforms the previously existing methods. It successfully achieves state-of-the-art results in all three datasets, namely 72.76% on the Okutama dataset, 92.56% on the MOD20 dataset, and 71.79% on the Drone Action dataset without pose-stream whereas 78.86% with pose-stream. For the Okutama dataset specifically, our Weighted Temporal

TABLE III
STATE-OF-THE-ART RESULTS ON THE MOD20 DATASET. BLUE
REPRESENTS THE PREVIOUS STATE-OF-THE-ART. RED DENOTES THE BEST
RESULTS.

| S.no. | Method | Backbone | Accuracy |
|---|---|---|---|
| Past | KRP-FS [27] [15] | VGG-f + motion-CNN | **74.00%** |
| Ours | BaseNet | Inception-v3 | 90.03% |
|  | **SWTF** | Weighted Temporal Fusion + Inception-v3 | **92.56%** |

TABLE IV
COMPARISON WITH THE STATE-OF-THE-ART RESULTS ON THE
DRONEACTION DATASET. BLUE REPRESENTS THE PREVIOUS
STATE-OF-THE-ART. RED DENOTES THE BEST RESULTS.

| S.no. | Method | Backbone | Accuracy |
|---|---|---|---|
| Past | HLPF [28] [18] | NTraj+ descriptors | 64.36% |
|  | PCNN [29] [18] | 'VGG-f' + Action Tubes | **75.92%** |
| Ours | BaseNet | Inception-v3 | 62.79% |
|  | **SWTF** | Weighted Temporal Fusion + Inception-v3 | 71.79% |
|  | **SWTF+Pose-Stream** | Weighted Temporal Fusion + Inception-v3+Pose-Stream | **78.86%** |

Fusion module with RoiAlign leads the network to focus on the keypoints where the action is currently taking place, and ignores the background noise, as opposed to the previously used 3D CNNs in [16]. It is also computationally less expensive, being a single stream network as compared to the approaches used in the previous works for MOD20 and Okutama dataset evaluation.

Jhuang *et al.* [28] uses the HLPF approach which focuses on temporal and spatial information but ignores the additional data of the objects or props used in performing the action. Consequently, Cheron *et al.* [29] P-CNN which uses the two-stream network to process RGB patches and flow patches is able to surpass HLPF results. With our simple CNN-LSTM model that is decently able to distinguish between similar classes, to get results using pose data and computationally cheap temporal segment network to process ROI cropped regions, our model is able to produce better results in a shorter amount of time.

CONCLUSION

In this study, we propose an SWTF network consisting of the Sparse Weighted Temporal Fusion module, which significantly improves our basenet's performance without significantly increasing its computational cost. We have introduced a novel method of fusing the concepts of temporal segment networks, weighted temporal fusion, and convolutional neural networks in order to determine the activity being performed in drone-camera-collected videos. Significant growth is observed in the Okutama-Action dataset, which can be extremely useful for drone-based activity recognition tasks at extremely high altitudes, such as crowd analysis and video surveillance. Our model achieves state-of-the-art performance on the challenging outdoor scenes depicted in the MOD20 and Drone-Action datasets, despite being less complex than alternative approaches. Deploying such a model on a suitable device could be increasingly advantageous.

REFERENCES

[1] N. Ikizler and D. Forsyth, "Searching video for complex activities with finite state models," 06 2007.
[2] L. Lo Presti and M. La Cascia, "3d skeleton-based human action classification," *Pattern Recogn.*, vol. 53, no. C, p. 130–147, May 2016. [Online]. Available: https://doi.org/10.1016/j.patcog.2015.11.019
[3] F. Lv and R. Nevatia, "Single view human action recognition using key pose matching and viterbi path searching," 06 2007.
[4] C. Zhang, Y. Zou, G. Chen, and L. Gan, "PAN: towards fast action recognition via learning persistence of appearance," *CoRR*, vol. abs/2008.03462, 2020. [Online]. Available: https://arxiv.org/abs/2008.03462
[5] J. Lin, C. Gan, and S. Han, "Tsm: Temporal shift module for efficient video understanding," in *Proceedings of the IEEE/CVF International Conference on Computer Vision (ICCV)*, October 2019.
[6] L. Wang, Y. Xiong, Z. Wang, and Y. Qiao, "Towards good practices for very deep two-stream convnets," 2015.
[7] Z. Li, E. Gavves, M. Jain, and C. G. M. Snoek, "Videolstm convolves, attends and flows for action recognition," 2016.
[8] D. Tran, H. Wang, L. Torresani, and M. Feiszli, "Video classification with channel-separated convolutional networks," in *Proceedings of the IEEE/CVF International Conference on Computer Vision (ICCV)*, October 2019.
[9] L. Wang, Y. Xiong, Z. Wang, Y. Qiao, D. Lin, X. Tang, and L. V. Gool, "Temporal segment networks: Towards good practices for deep action recognition," *CoRR*, vol. abs/1608.00859, 2016. [Online]. Available: http://arxiv.org/abs/1608.00859
[10] B. Zhou, A. Andonian, A. Oliva, and A. Torralba, "Temporal relational reasoning in videos," 2018.
[11] X. Song, C. Lan, W. Zeng, J. Xing, X. Sun, and J. Yang, "Tem- poral–spatial mapping for action recognition," *IEEE Transactions on Circuits and Systems for Video Technology*, vol. 30, no. 3, pp. 748–759,2020.
[12] Y. Li, W. Li, V. Mahadevan, and N. Vasconcelos, "Vlad3: Encoding dynamics of deep features for action recognition," in *2016 IEEE Conference on Computer Vision and Pattern Recognition (CVPR)*, 2016, pp. 1951–1960.
[13] R. Girdhar, D. Ramanan, A. Gupta, J. Sivic, and B. Russell, "Actionvlad: Learning spatio-temporal aggregation for action classification," 2017.
[14] G. Chéron, I. Laptev, and C. Schmid, "P-cnn: Pose-based cnn features for action recognition," 2015.
[15] A. Perera, Y. Law, T. Ogunwa, and J. Chahl, "A multiviewpoint outdoor dataset for human action recognition," *IEEE Transactions on Human-Machine Systems*, vol. PP, pp. 1–9, 02 2020.
[16] A. M. Algamdi, V. S. Silva, and C.-T. Li, "Dronecaps : recognition of human actions in drone videos using capsule networks with binary volume comparisons," in *27th IEEE International Conference on Image Processing*. IEEE, 2020. [Online]. Available: http://wrap.warwick.ac.uk/141611/
[17] M. Barekatain, M. Martı́, H. Shih, S. Murray, K. Nakayama, Y. Matsuo, and H. Prendinger, "Okutama-action: An aerial view video dataset for concurrent human action detection," *CoRR*, vol. abs/1706.03038, 2017. [Online]. Available: http://arxiv.org/abs/1706.03038
[18] A. G. Perera, Y. W. Law, and J. Chahl, "Drone-action: An outdoor recorded drone video dataset for action recognition," *Drones*, vol. 3, no. 4, 2019. [Online]. Available: https://www.mdpi.com/2504-446X/3/4/82
[19] R. Gao, B. Xiong, and K. Grauman, "Im2flow: Motion hallucination from static images for action recognition," *CoRR*, vol. abs/1712.04109, 2017. [Online]. Available: http://arxiv.org/abs/1712.04109
[20] C. Szegedy, V. Vanhoucke, S. Ioffe, J. Shlens, and Z. Wojna, "Rethinking the inception architecture for computer vision," *CoRR*, vol. abs/1512.00567, 2015.

[21] J. Deng, W. Dong, R. Socher, L.-J. Li, K. Li, and L. Fei-Fei, "Imagenet: A large-scale hierarchical image database," in *2009 IEEE Conference on Computer Vision and Pattern Recognition*, 2009, pp. 248–255.
[22] K. Simonyan and A. Zisserman, "Two-stream convolutional networks for action recognition in videos," 2014.
[23] F. Yang, S. Sakti, Y. Wu, and S. Nakamura, "A framework for knowing who is doing what in aerial surveillance videos," *IEEE Access*, vol. PP, pp. 1–1, 07 2019.
[24] M. Zolfaghari, K. Singh, and T. Brox, "Eco: Efficient convolutional network for online video understanding," in *Proceedings of the European Conference on Computer Vision (ECCV)*, September 2018.
[25] J. Carreira and A. Zisserman, "Quo vadis, action recognition? a new model and the kinetics dataset," in *2017 IEEE Conference on Computer Vision and Pattern Recognition (CVPR)*, 2017, pp. 4724–4733.
[26] P. ZHang, P. Wei, and S. Han, "CapsNets algorithm," *Journal of Physics: Conference Series*, vol. 1544, p. 012030, may 2020. [Online]. Available: https://doi.org/10.1088/1742-6596/1544/1/012030
[27] A. Cherian, S. Sra, S. Gould, and R. Hartley, "Non-linear temporal subspace representations for activity recognition," *CoRR*, vol. abs/1803.11064, 2018. [Online]. Available: http://arxiv.org/abs/1803.11064
[28] H. Jhuang, J. Gall, S. Zuffi, C. Schmid, and M. Black, "Towards understanding action recognition," 12 2013, pp. 3192–3199.
[29] G. Chéron, I. Laptev, and C. Schmid, "P-cnn: Pose-based cnn features for action recognition," in *2015 IEEE International Conference on Computer Vision (ICCV)*, 2015, pp. 3218–3226.